\documentclass[letterpaper, 10 pt, conference]{ieeeconf}  

\IEEEoverridecommandlockouts                              

\usepackage{makecell}
\usepackage{graphicx}
\usepackage{xcolor}
\usepackage{soul}

\title{\LARGE \bf
Spiking Neural Network on Neuromorphic Hardware for Energy-Efficient Unidimensional SLAM
}

\author{Guangzhi Tang, Arpit Shah, and Konstantinos P. Michmizos
\thanks{*This work is supported by Intel's INRC Grant Award}
\thanks{GT, AS and KM are with the Computational Brain Lab, Department of Computer Science, Rutgers University, New Jersey, USA.
        {\tt\small konstantinos.michmizos@cs.rutgers.edu}}%
}

\begin{document}

\maketitle
\thispagestyle{empty}
\pagestyle{empty}

\begin{abstract}

Energy-efficient simultaneous localization and mapping (SLAM) is crucial for mobile robots exploring unknown environments. The mammalian brain solves SLAM via a network of specialized neurons, exhibiting asynchronous computations and event-based communications, with very low energy consumption. We propose a brain-inspired spiking neural network (SNN) architecture that solves the unidimensional SLAM by introducing spike-based reference frame transformation, visual likelihood computation, and Bayesian inference. We integrated our neuromorphic algorithm to Intel's Loihi neuromorphic processor, a non-Von Neumann hardware that mimics the brain's computing paradigms. We performed comparative analyses for accuracy and energy-efficiency between our neuromorphic approach and the GMapping algorithm, which is widely used in small environments. Our Loihi-based SNN architecture consumes 100 times less energy than GMapping run on a CPU while having comparable accuracy in head direction localization and map-generation. These results pave the way for scaling our approach towards active-SLAM alternative solutions for Loihi-controlled autonomous robots.

\end{abstract}

\section{Introduction}

Localization, knowing one's pose, and mapping, knowing the positions of surrounding landmarks, are essential skills for both humans and robots as they navigate in unknown environments. The main challenge is to produce accurate estimates from noisy, error-prone cues, with robustness, efficiency, and adaptivity. Graph-based \cite{kummerle2011g,dellaert2010subgraph} and filter-based approaches \cite{grisetti2007improved,strom2011occupancy} have solved the simultaneous localization and mapping (SLAM) problem by either optimizing a constrained graph or performing recursive Bayesian estimation. As they are tackling SLAM in a growing number of real-world applications, these approaches face increasing challenges for minimizing energy consumption.

Interestingly, efficient and highly accurate localization and mapping are "effortless" characteristics of mammalian brains \cite{poulter2018neurobiology}. Over the last few decades, a number of specialized neurons, including border cells, head direction cells, place cells, grid cells, and speed cells, have been found to be part of a brain network that solves localization and mapping \cite{grieves2017representation} in an energy-efficient manner \cite{sengupta2014power}.

Large-scale neuromorphic processors \cite{davies2018loihi, merolla2014million, schemmel2010wafer, furber2014spinnaker} have been proposed as a non-Von Neumann alternative to traditional computing hardware. These processors offer asynchronous event-based parallelism and relatively efficient solutions to many mobile robot applications \cite{mei2005case,blum2017neuromorphic,hwu2017self}. Particularly, Intel's Loihi processor \cite{davies2018loihi} supports on-chip synaptic learning, multilayer dendritic trees, and other brain-inspired components such as synaptic delays, homeostasis, and reward-based learning.

To leverage the disruptive potential of neuromorphic computing, we need to develop new algorithms that call for a bottom-up rethinking of our already developed solutions. Neuromorphic processors are designed to run Spiking Neural Networks (SNN), a specialized brain-inspired architecture where simulated neurons emulate the learning and computing principles of their biological counterparts. SNNs can introduce an orthogonal dimension to neural processing by adhering to the structure of the biological networks associated with the targeted tasks. Specifically, the brain's spatial navigation and sensorimotor systems have inspired the design of SNNs that solved a number of problems in robotics \cite{bing2018end,bing2018survey,hwu2018adaptive} Of particular interest for this study is an SNN inspired by the brain's navigational system that enables a mobile robot to correct its estimate of pose and map of a simple environment, by periodically using a ground-truth signal \cite{kreiser2018pose}.

In this paper, we present a biologically constrained SNN architecture which solves the unidimensional SLAM problem on Loihi, without depending on the external ground truth information. To do so, our proposed model determines the robot's heading via spike-based recursive Bayesian inference of multisensory cues, namely visual and odometry information. We validated our implementation in both real-world and simulated environments, by comparing with the GMapping algorithm \cite{grisetti2007improved}. The SNN generated representations of the robot's heading and mapped the environment with comparable performance to the baseline while consuming less than 1\% of dynamic power.

\section{Method}

We developed a recursive SNN that suggests a cue-integration connectome performing head direction localization and mapping, and we integrated the network to Loihi. Inspired by the spatial navigation system found in the mammalian brain, the head direction and border cells in our network exhibited biologically realistic activity \cite{grieves2017representation}. Our model had intrinsic asynchronous parallelism by incorporating spiking neurons, multi-compartmental dendritic trees, and plastic synapses, all of which are supported by Loihi.

\begin{figure}
\vspace{5.2pt}
\centering
\includegraphics[scale=1.0]{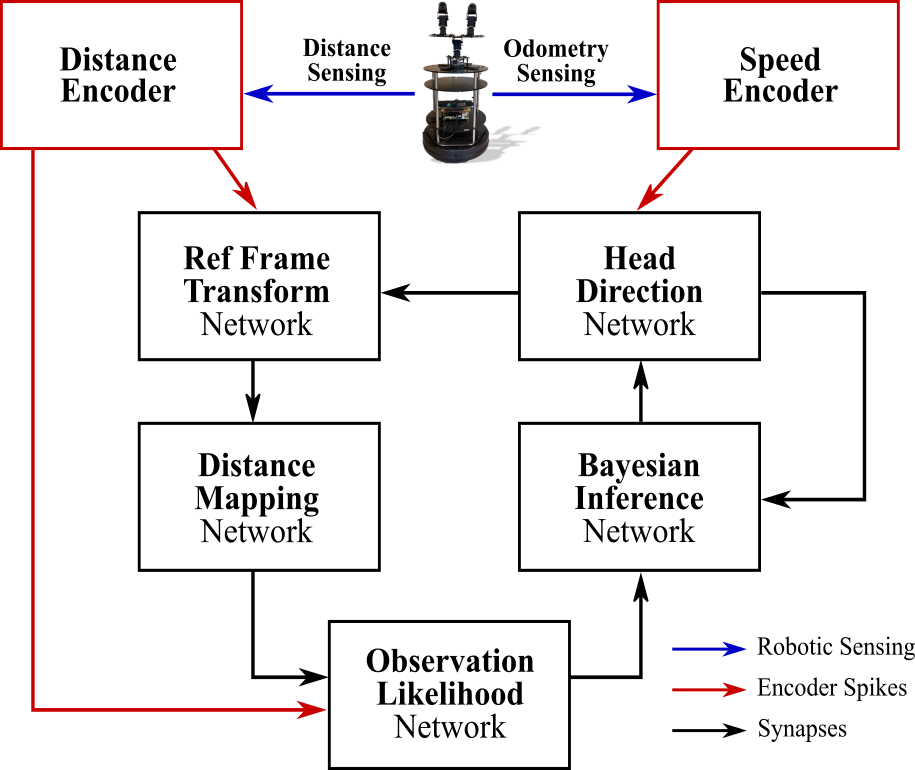}
\caption{Structure of the SNN architecture. Each block is a sub-network.}
\label{fig: overall}
\end{figure}

Our model had 2 sensory spike rate-encoders and 5 sub-networks (Fig. \ref{fig: overall}). The odometry sensor and the RGB-Depth camera signals drove the neural activity of speed cells and sensory neurons encoding the angular speed and the distance to the nearest object, respectively. The Head Direction (HD) network received the input from the speed cells and represented the heading of the robot. The Reference Frame Transformation (RFT) network received the egocentric input from sensory neurons and generated allocentric distance representation in the world reference frame, as defined by the HD network. The Distance Mapping (DM) network learned the allocentric observations from the RFT network and formed the map of the robot's surrounding environment. The Observation Likelihood (OL) network used the map from the DM network to compute the observation likelihood distribution of the robot's heading based on the egocentric observation from sensory neurons. The Bayesian Inference (BI) network produced a near-optimal posterior of the robot's heading and corrected the heading representation within the HD network. To do so, the BI network used the observation likelihood from the OL network and the odometry likelihood from the HD network. Each one of the networks is briefly described below. 

\subsection{Head Direction Network}

Head direction cells in the HD network changed their spiking activity according to the robot's heading, as follows. The HD network comprised of 75 neurons, each having a 5-degree resolution. We used the Continuous Attractor Model \cite{wu2016continuous} to integrate angular speed and form a stable representation of the robot's heading (Fig. \ref{fig: SNN}a). The HD attractor state shifted either clockwise or counter-clockwise, depending on the robot's rotation, with the help of transition neurons. There were two populations of such neurons to represent the two possible directions of rotation. Each transition neuron had a dendritic tree with one dendrite receiving spikes from the speed cell and the other from its corresponding head direction cell. The neuron fired when both dendrites were activated, thereby changing the HD attractor state.

\subsection{Reference Frame Transformation Network}

\begin{figure*}
\vspace{5.2pt}
\centering
\includegraphics[scale=1.0]{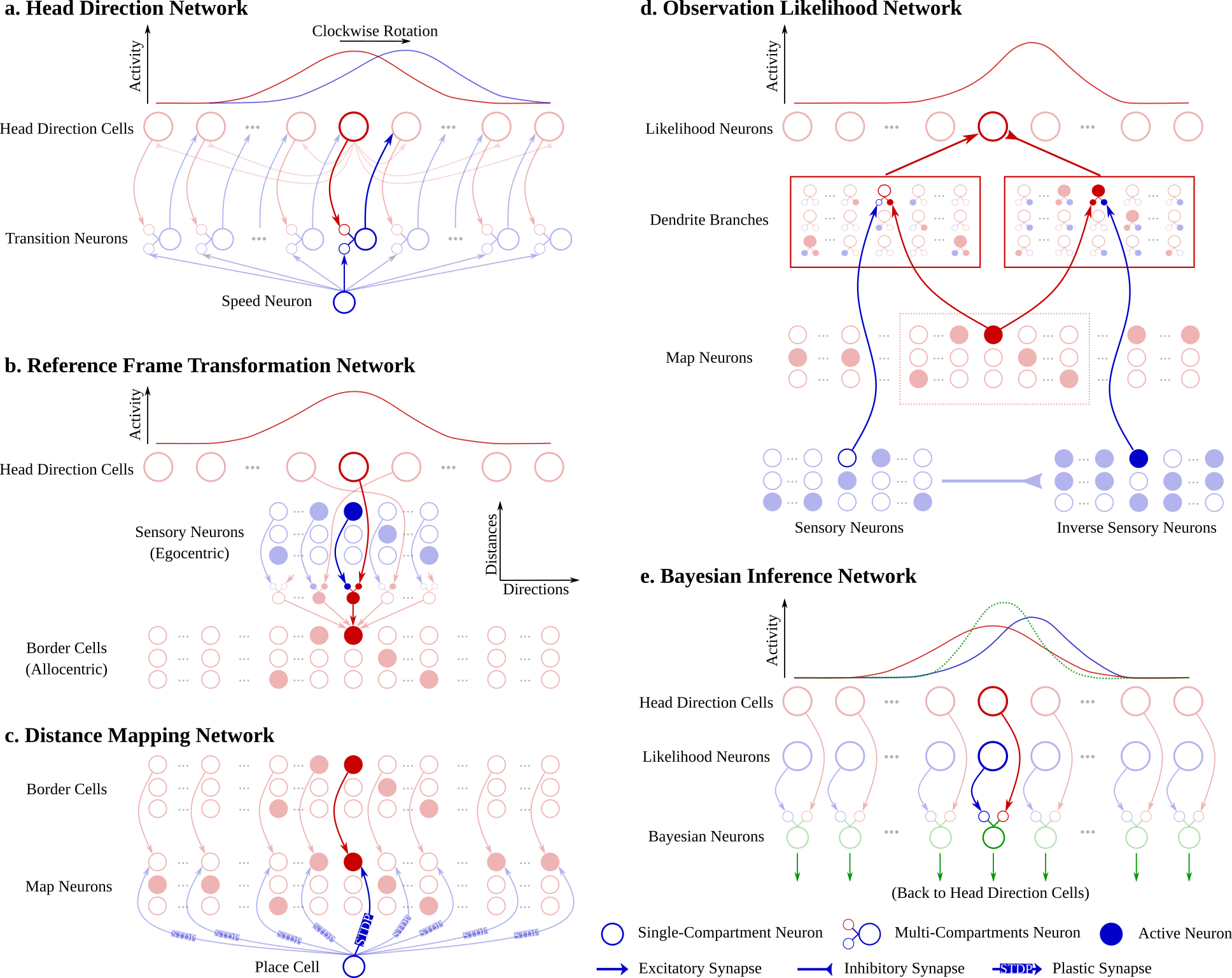}
\caption{Connectome of each sub-network in the SNN architecture: a) Head direction network with one group of transition neurons; b) Reference frame transformation network with connections to a single border cell; c) Distance mapping network with connections to map neurons representing one of the three distances; d) Observation likelihood network with connections to one likelihood neuron; e) Bayesian inference network.}
\label{fig: SNN}
\end{figure*}

Border cells in the RFT network represented distance observations in the world reference frame (Fig. \ref{fig: SNN}b). The sensory neurons represented discretized distances between the observable objects and the robot, in an egocentric manner. Similarly to our previous work \cite{tang2018gridbot}, the RFT network used the HD activity to create an allocentric representation of the surrounding environment, therefore translating from egocentric observations to mapping. Other spike-based methods exist that perform reference frame transformation \cite{schneegans2012neural,blum2017neuromorphic,bicanski2018neural}. 

To perform the transformation in the RFT network, we leveraged the concurrent activity of sensory neurons and HD cells, as follows. Sensory neurons encoded the depth signal at the robot's heading, as represented by HD cells. We created a group of border cells with the same preferred headings as the HD cells, allowing the border cells and HD cells to be on the same reference frame and have a one-to-one correspondence on preferred headings. Each border cell had a dendritic tree receiving spikes from HD cells and sensory neurons. A border cell fired maximally when the HD cells and sensory neurons connected to its dendritic tree were activated at the same time.

\subsection{Distance Mapping Network}

The spike activity of map neurons in the DM network represented the mapping of the robot's surrounding environment (Fig. \ref{fig: SNN}c). The map was stored in the synapses between a single place cell and all map neurons, using an unsupervised, Hebbian-type rule. When the map (post-synaptic) neuron fired, the synaptic weight, $w$, increased proportionally to the trace of the place cell's (pre-synaptic) spikes, as follows:
$$
\delta w = A*x_1*y_0 - B*u_k, \eqno{(1)}
$$
where the trace $x_1$ was the convolution of the pre-synaptic neuron's spikes with a decaying exponential function; $y_0$ changes from 0 to 1 whenever a post-synaptic neuron fires; and $u_k$ is a decay factor which changed from 0 to 1 every k time-steps and prevented overlearning in synapses with inconsistent pre-synaptic activity. That way, the network learned only the obstacles in the map that were observed with high certainty. During learning, map neurons were activated by border cells, and a single place cell was activated by the location of the robot. A winner-take-all (WTA) mechanism was implemented as an inhibition of the nearby map neurons and ensured that a single map neuron would be active at each location. 

\subsection{Observation Likelihood Network}

Likelihood neurons in the OL network changed their spike activity based on the encoded distances and formed an observation likelihood distribution (Fig. \ref{fig: SNN}d). The network encoded the likelihoods of different headings based on the observed distance pattern. The distribution was multimodal when multiple similar distance patterns existed in the environment. This enabled the robot to estimate its heading without reference to its odometry sensor. The OL network is a spike-based alternative to the previously proposed scan matching methods \cite{olson2009real, olson2015m3rsm}, which compute observation likelihoods based on visual observations.

To generate the likelihood activity, we computed similarities between the depth signal and the map, by employing asynchronous dendritic processing, as follows. Synaptic connections from map neurons to OL neurons formed spatial windows in the learned representations of the environment. Since this pattern comparison considered only the excited neurons between the observation and the map, it could generate wrong likelihoods. To overcome this, we used a group of inverse sensory neurons to compute the similarity of the inverse distance pattern with the map. This second dendritic branch inhibited the likelihood, since it represented the non-active part of the environment. These two branches of the OL neurons, increased the contrast in inferring the heading.

\subsection{Bayesian Inference Network}

Bayesian neurons in the BI network generated a posterior distribution from the likelihood functions (Fig. \ref{fig: SNN}e), as defined in Equation 2:
$$
p(s|d,o) \propto p(d|s)p(o|s)p(s), \eqno{(2)}
$$
 where $s$ is the heading of the robot, $d$ is the distance observed, $o$ is the odometry sensing. With a flat prior $p(s)$, the posterior distribution over the robot's heading is proportional to the product of two likelihood functions, $p(d|s)$ and $p(o|s)$, through Bayes' theorem.

It is known that multiplying two Gaussian distributions produces another Gaussian distribution. This property enabled us to use dendritic trees to estimate the posterior distribution from likelihood distributions represented by the OL network and the HD network. Specifically, each Bayesian neuron had two dendritic compartments connected with its corresponding OL neuron and HD cell. The \textit{PASS} dendritic operation on Loihi integrated the OL neuron voltage into the Bayesian neuron voltage when the HD cell spiked. Through this operation, the Bayesian neuron estimated the product of activities from the OL neuron and the HD cell.

\begin{figure}
\vspace{5.2pt}
\centering
\includegraphics[scale=1.0]{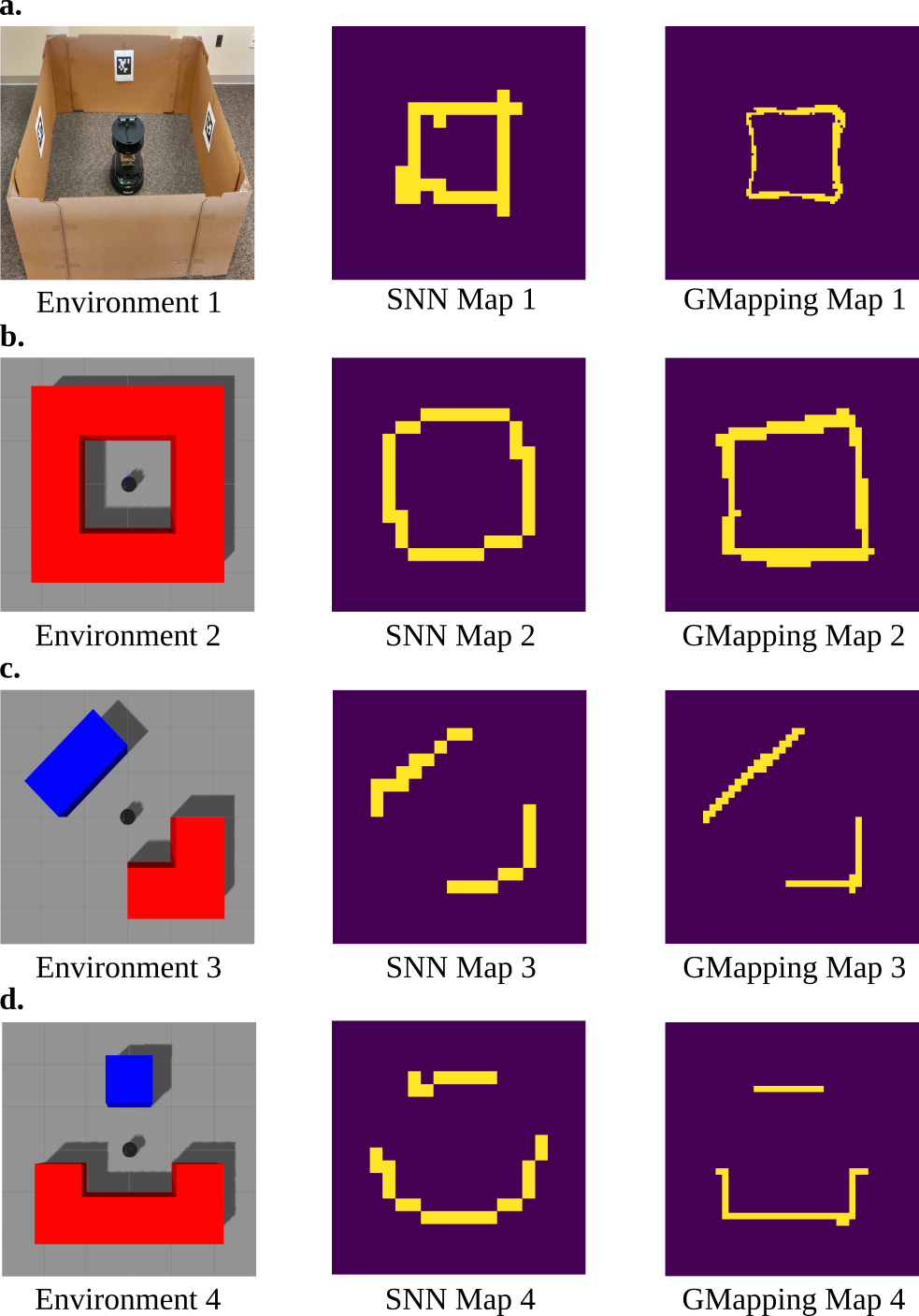}
\caption{(left column) Experimental environments; (middle column) Learned maps as represented by map neurons in our SNN architecture; (right column) Learned maps generated by the GMapping algorithm, with the lowest resolution that gave a stable solution.}
\label{fig: map}
\end{figure}

\subsection{Neuromorphic Realization in Loihi}

We implemented our SNN architecture in one Loihi research chip. With a mesh layout, Loihi supports 128 neuromorphic cores with 1,024 compartments (primitive spiking neural units) in each core. Overall, a single chip provides up to 128k neurons and 128M synapses for building large-scale SNNs \cite{davies2018loihi}. Our SNN architecture used 15,162 compartments and 31,935 synapses distributed over 82 neuromorphic cores, slightly more than ten percent of the resources in a single Loihi research chip. When encoding the input from the distance observation, the encoder transformed all values to 3 discrete distance levels. Additionally, all neurons with HD receptive fields had a resolution of 5 degrees. For example, each sensory neuron encoded a single distance level for representing objects observable within 5 degrees.

\section{Experiment and Results}

\subsection{Experimental Setup}

We used a mobile robot equipped with an RGB-Depth camera, in both the real-world and Gazebo simulator, for validating our method. During all experiments, the robot rotated for 120 seconds with only angular velocity commands. We created 1 real-world and 3 simulated environments (Fig. \ref{fig: map}). Environments 1 and 2 provided continuous borders with environment 2 simulating the real-world environment. We also considered more simulated scenarios where non-continuous objects (Environments 3 and 4) left gaps between themselves. In the simulated environments, we retrieved the ground truth of the robot's heading directly from Gazebo model states. In the real-world environment, we used the AprilTag detection system \cite{olson2011apriltag} and 4 AprilTag tags to determine sufficiently the ground truth values. 

\begin{figure}[!b]
\vspace{5.2pt}
\centering
\includegraphics[scale=1.0]{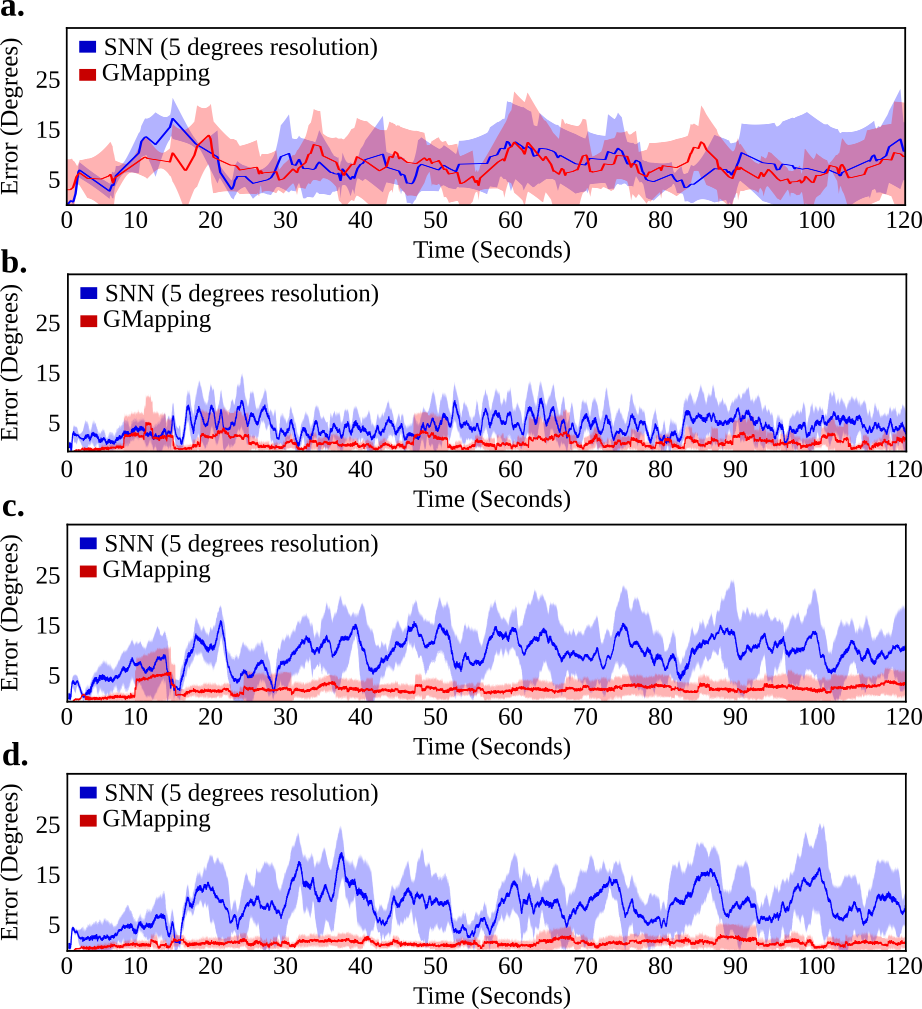}
\caption{Mean and STD of localization error over 5 experiments for both methods in a) Environment 1 (real-world environment), b) Environment 2, c) Environment 3, and d) Environment 4.}
\label{fig: arrcuracy}
\end{figure}

\subsection{The Baseline Method}
We chose the GMapping algorithm \cite{grisetti2007improved} as the baseline method solving the same unidimensional SLAM problem. To equally compare GMapping with our method, we limited GMapping to the lowest resolution that gave stable results. For the real-world environment, GMapping built the map using a resolution of 0.04 meters and did scan-matching using all distance data from each update with a minimum score parameter of 700. For the simulated environments, GMapping built maps using a resolution of 0.1 meters and did scan-matching using 15 evenly distributed distance observations with a minimum score parameter of 10.

\subsection{Localization and Mapping}

We compared the heading from the HD cells with the ground truth values (Fig. \ref{fig: arrcuracy}). We conducted 5 experiments in the 4 environments and estimated the average error of headings to less than 15 degrees, for both our method and GMapping. Given the 5 degrees resolution of the HD cells, the error was in practice a 1 to 3 neuron-drift in the attractor model, which had up to 10 active neurons. We observed a higher variance in the errors for environments 3 and 4, which was due to the free space between the objects and the instability in correcting the activity of the attractor model. Indeed, when there was no object observed, the error increased temporarily until an object was within the range of the visual observation. Similarly to any filter-based approach on SLAM, as soon as an object was detected, there was a sharp correction resulting in error decrease (Figs. \ref{fig: arrcuracy}c and d). 

\begin{figure}
\vspace{5.2pt}
\centering
\includegraphics[scale=1.0]{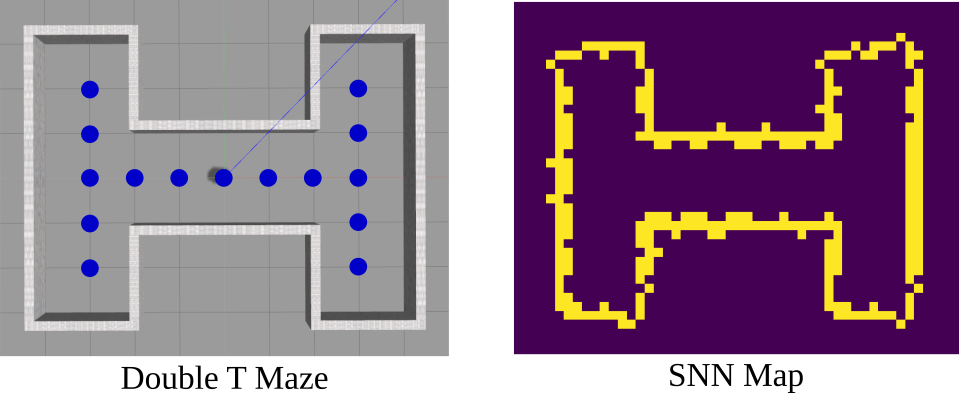}
\caption{Learned double T maze environment to demonstrate scalability using multiple unidimensional maps. A single place cell in the DM network corresponded to a location in the maze (blue dots). The learned 2D map was constructed by superimposing the maps from all place cells.}
\label{fig: bigmap}
\end{figure}

We decoded the activity of the map neurons into a 20x20 gridmap representing a 4mx4m environment. Environments 1 and 2 had a square shape, and the maps generated by the SNN (Fig. \ref{fig: map}a,b) successfully captured the repetitive distance pattern at the corners. Environments 3 and 4 had two objects with different shapes. The maps learned by our method (Fig \ref{fig: map}c,d) reflected the differences between the two objects as perceived by the robot. We further show how our SNN can scale to map a 2D environment by using more than one place cell in the DM network (Fig. \ref{fig: bigmap}).

\subsection{Observation Likelihood and Bayesian Inference}

\begin{figure}
\vspace{5.2pt}
\centering
\includegraphics[scale=1.0]{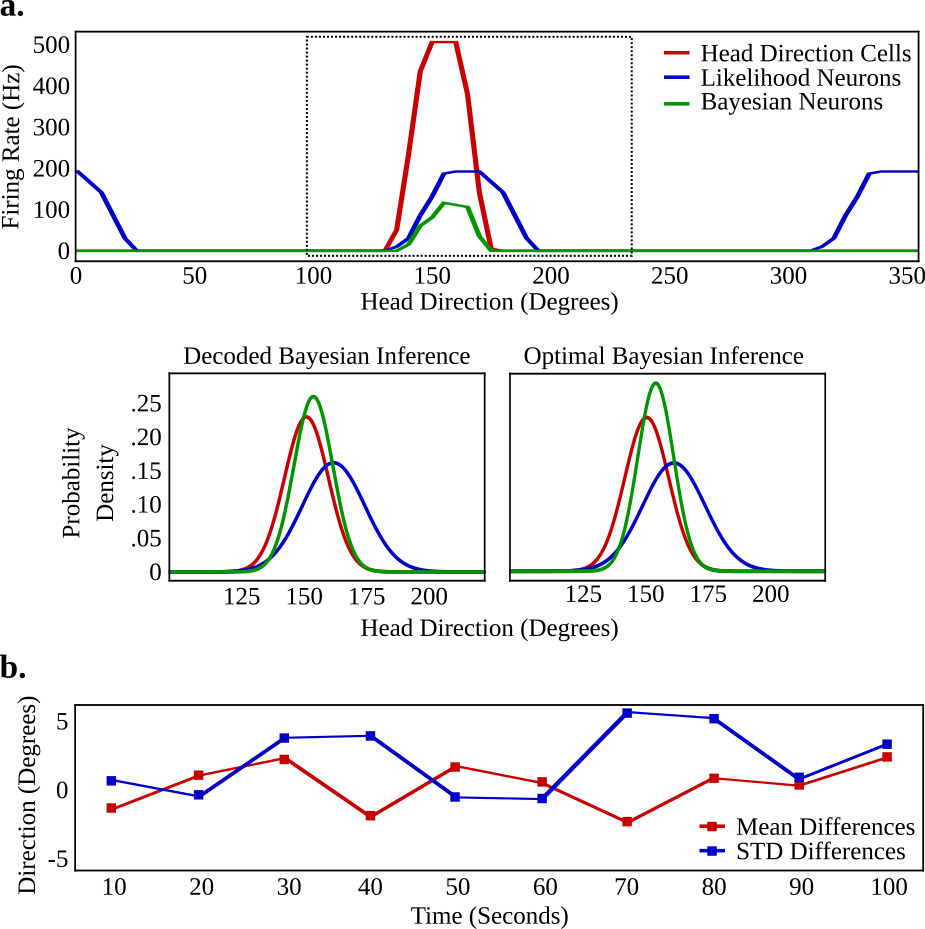}
\caption{Spike-based Bayesian inference. a) (upper panel) Neuronal activities within the BI network for Environment 4 and (bottom panel) comparison between the decoded and the optimal Bayesian inference results. b) Mean and STD differences between the decoded and the optimal posterior distribution during a single run in Environment 4.}
\label{fig: likelihood}
\end{figure}

The activity of OL neurons captured the distinctive patterns in the learned environment. For instance, firing rates of OL neurons in Fig. \ref{fig: likelihood}a formed a bimodal distribution representing two possible headings due to the repetitive objects in Environment 4. We evaluated the activity of the Bayesian neurons by decoding the spikes from HD cells and OL neurons within a range of head directions into two Gaussian distributions, $N_1(\mu_1, \sigma^2_1)$ (red) and $N_2(\mu_2, \sigma^2_2)$ (blue) respectively. Equations (3) and (4) give the optimal posterior distribution $N_3(\mu_3, \sigma^2_3)$ (green) from these two likelihood distributions (Fig. \ref{fig: likelihood}a):
$$
\mu_3 = \frac{\sigma^2_2}{\sigma^2_1+\sigma^2_2}\mu_1 + \frac{\sigma^2_1}{\sigma^2_1+\sigma^2_2}\mu_2 \eqno{(3)}
$$
$$
\sigma^2_3 = \frac{1}{\frac{1}{\sigma^2_1} + \frac{1}{\sigma^2_2}} \eqno{(4)}
$$
We also computed the differences of the means and the standard deviations (STDs) between the decoded posterior distribution from Bayesian neurons and the optimal posterior distribution during runtime (Fig. \ref{fig: likelihood}b). During the experiments, the difference of the mean and STD was always less than 5 degrees, which is, in fact, the resolution of the head direction in our SNN. The transient increase in the STD differences in Fig \ref{fig: likelihood}b was caused by the small resolution, constrained to 2 or 3 neurons, for representing the posterior distribution. Overall, the BI network generated near-optimal posterior distribution by performing spike-based Bayesian inference.

\begin{figure}[!b]
\centering
\includegraphics[scale=1.0]{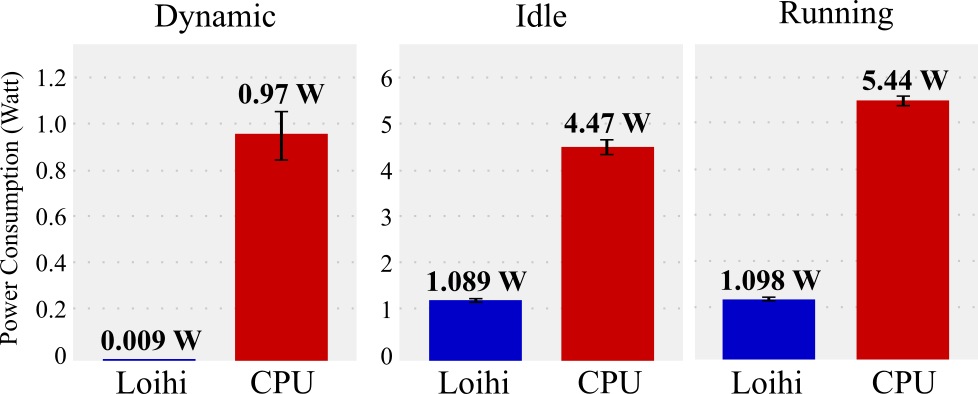}
\caption{Power consumption of our SNN architecture ran on Loihi and GMapping ran on CPU, solving the same unidimensional SLAM problem.}
\label{fig: power}
\end{figure}

\subsection{Energy Efficiency}
Our Loihi-run SNN was two orders of magnitude more energy efficient compared to the CPU-run GMapping solving the same unidimensional SLAM problem (Fig. \ref{fig: power}). We compared the power consumption of the SNN on a Nahuku board, an 8-chip Loihi system, with that of GMapping on an Intel i7-4850HQ CPU. To measure the idle power on Loihi, we set all compartments to non-updating state in multiple 10,000 time-step runs. The idle CPU power was measured by running only the operating system for ten minutes. The running power for both methods was averaged over six such experiments.

Similarly to GMapping, our SNN architecture performed real-time data processing by only using 0.3 seconds for execution per wall-clock second, on average. This allowed us to compare the Loihi power consumption and the CPU power consumption against the same wall-clock time of the running robot (Fig. \ref{fig: power}). We computed the dynamic power of each method by subtracting the idle power from the running power. An 8-chip Loihi board uses 4 times less power compared to a quad-core CPU in the idle state and our SNN running on Loihi was 100 times more energy efficient compared to GMapping running on a CPU in terms of dynamic power consumption. Since Loihi is at an early development stage, the power consumption, especially the idle power consumption, can be lowered further to 0.031 Watts in a more customized system \cite{blouw2018benchmarking} compared to the Nahuku board we utilized.

\section{Discussion and Conclusion}

In this paper, we showed that an SNN architecture inspired by the brain's spatial navigation system and run on a neuromorphic processor can have similar accuracy and much lower power consumption, compared to a widely used method for solving the unidimensional SLAM problem.  Although the error in the sparse environments was larger than GMapping, our proof-of-concept results can be improved by increasing the resolution or the stability of the HD network, to further demonstrate the validity of our proposed method as similarly accurate and much more efficient in terms of power consumption SLAM method. Similar to other solutions running on neuromorphic processors addressing speech recognition \cite{blouw2018benchmarking} and image processing \cite{davies2018loihi}, our approach currently yields results that are only comparable to the state-of-the-art methods that have been well-tuned to run on traditional Von Neumann CPUs. 

For applications such as planetary exploration and disaster rescue, where robots have limited recharging capabilities, minimizing energy consumption is crucial. Our proposed neuromorphic approach provides an energy efficient solution to the SLAM problem, which accounts for a large portion of the computational cost and its energy consumption.

Overall, this work points to the real-time neuromorphic control of robots as a strong alternative, complementing widely used solutions for foundational robotic problems. Although it probably requires a lot more small insights before it can scale to outperform a highly developed technology, the fact that our Loihi-run SNN compares in accuracy to a mainstream method while offering unparalleled energy efficiency, is an indication that end-to-end neuromorphic solutions for fully autonomous systems is a direction worth exploring.

\bibliographystyle{IEEEtran}
\bibliography{myref}

\end{document}